\documentclass[runningheads]{llncs}
\usepackage{graphicx}
\usepackage{cite}
\usepackage{bbding}
\usepackage{multirow}
\usepackage{booktabs}
\usepackage[misc]{ifsym} 
\usepackage[colorlinks]{hyperref}

\begin{document}
\title{Unified Attentional Generative Adversarial Network for Brain Tumor Segmentation From Multimodal Unpaired Images}
\author{Wenguang Yuan\inst{1} \and 
	Jia Wei\inst{1,}\textsuperscript{\Letter} \and 
	Jiabing Wang\inst{1} \and 
	Qianli Ma\inst{1} \and 
	Tolga Tasdizen\inst{2}
}
\authorrunning{W. Yuan et al.}
\titlerunning{UAGAN for Brain Tumor Segmentation}
\institute{\textsuperscript{1}School of Computer Science and Engineering, South China University of Technology, Guangzhou, China\\
	\textsuperscript{2}Scientific Computing and Imaging Institute, University of Utah, Salt Lake City, USA\\
}
\maketitle 

\begin{abstract}
In medical applications, the same anatomical structures may be observed in multiple modalities despite the different image characteristics. Currently, most deep models for multimodal segmentation rely on paired registered images. However, multimodal paired registered images are difficult to obtain in many cases. Therefore, developing a model that can segment the target objects from different modalities with unpaired images is significant for many clinical applications. In this work, we propose a novel two-stream translation and segmentation unified attentional generative adversarial network (UAGAN), which can perform any-to-any image modality translation and segment the target objects simultaneously in the case where two or more modalities are available. The translation stream is used to capture modality-invariant features of the target anatomical structures. In addition, to focus on segmentation-related features, we add attentional blocks to extract valuable features from the translation stream. Experiments on three-modality brain tumor segmentation indicate that UAGAN outperforms the existing methods in most cases. 
\keywords{Brain Tumor Segmentation \and Image Translation \and Unpaired Images \and Adversarial Learning.}
\end{abstract}
\section{Introduction}
Gliomas, the most common primary central nervous system malignancies, consist of various subregions~\cite{bakas2017advancing}. In medical practice, multimodal images can provide different biological information mapping tumor-induced tissue change, such as postcontrast T1-weighted (T1Gd), T2-weighted (T2), and T2 Fluid Attenuated Inversion Recovery (FLAIR) volumes~\cite{menze2015multimodal}. The same anatomical structures may be observed in multiple modalities despite the different image characteristics. Currently, most deep models for multimodal segmentation rely on paired registered images~\cite{menze2015multimodal,tseng2017joint,nie2016fully}. However, multimodal paired registered images are difficult to obtain in many cases. Therefore, developing a model that can segment the target objects from different modalities with unpaired images is significant for many clinical applications.

Recently, the problem of multimodal segmentation has been extensively studied. Nie \emph{et al.}~\cite{nie2016fully} trained a network for three modalities and fused their high-layer features for the final segmentation. Tseng \emph{et al.}~\cite{tseng2017joint} proposed cross-modality convolution layers to better leverage multimodal information. However, these methods were limited because they required paired registered images. An alternative approach is to train different models for different modalities in a shared latent space by extracting a common representation from different modalities. Kuga \emph{et al.}~\cite{kuga2017multi} trained multimodal encoder-decoder networks with a shared representation. Valindria \emph{et al.}~\cite{valindria2018multi} extracted modality-independent features by sharing the last layers of the encoder. However, these methods were not unified and required more parameters because of the requirement of specific encoder-decoder architectures for each modality. To extract modality-invariant features efficiently, Xu \emph{et al.}~\cite{xu2018pad} represented a multimodal distillation module. Hu \emph{et al.}~\cite{hu2018squeeze} performed feature recalibration using SE (squeeze-and-excitation) blocks. Recently, adversarial learning has been regarded as an effective way to transfer knowledge across different image domains. Huo \emph{et al.}~\cite{huo2018adversarial} presented an end-to-end synthesis and segmentation network with unpaired MRI and CT images. To address limited scalability and robustness in translating among more than two domains, Choi \emph{et al.}~\cite{choi2018stargan} developed a scalable approach (StarGAN) that can perform image-to-image translation for multiple domains using an unified model.

During the adversarial learning of StarGAN, the generator learns to change the image characteristics and preserves the global features to fool the discriminator. Utilizing these global features may improve the performance of multimodal segmentation. Thus, in this work, we take translation as an auxiliary task to help segmentation and propose a two-stream translation and segmentation unified attentional generative adversarial network (UAGAN). In the translation stream, the discriminator is the same as StarGAN~\cite{choi2018stargan}, whereas the backbone of the generator is changed to U-net~\cite{ronneberger2015u} in order to better leverage the low-level and high-level features. In the segmentation stream, another U-net is adopted to share the last layers of the encoder in the translation stream. Because not all features extracted from the translation stream are useful for segmentation, we add attentional blocks to focus on the useful features for segmentation. Experiments on three-modality brain tumor segmentation indicate that UAGAN outperforms existing methods in most cases.
\section{UAGAN for Multimodal Segmentation}
Multimodal segmentation using a single model remains very challenging due to the different image characteristics of different modalities. A key challenge is to extract modality-invariant features. Previous multimodal segmentation methods have required paired \emph{n}-modality images. To address the limitation of requiring paired multimodal images, we present a two-stream UAGAN. The following subsections discuss the details of the proposed framework.
\subsection{Method Overview} \label{method overview}
\begin{figure}
	\includegraphics[width=\textwidth]{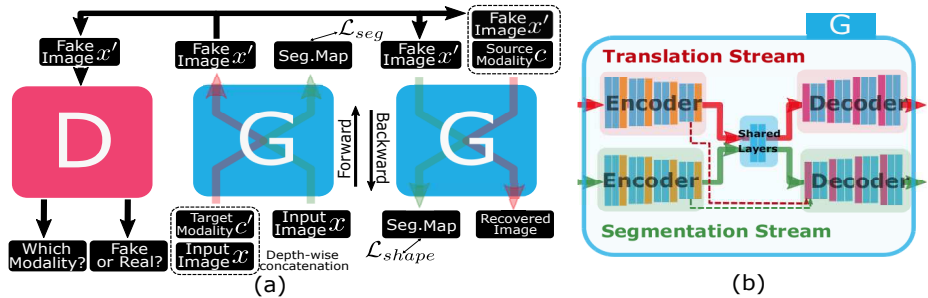}
	\caption{Overview of UAGAN. (a) The schematic illustration of training strategy .(b) The schematic illustration of $G$, where pink rectangles in decoders are attentional blocks, orange ones in encoders are max-pooling layers and blue ones are convolution blocks. Only lines (green for segmentation, red for translation) of the first attentional blocks in the segmentation stream are drawn.} \label{xnet2}
\end{figure}

Fig.~\ref{xnet2}a illustrates the training strategy of the proposed UAGAN and Fig.~\ref{xnet2}b shows the architecture of our UAGAN with the translation and segmentation streams. Both streams adopt the U-net architecture. Inspired by~\cite{valindria2018multi}, we adopt independent encoders and decoders but share the last layers of the encoders. We denote the network of the translation stream as $G_{trans}$ and the network of the segmentation stream as $G_{seg}$. The adversarial training strategy is similar to StarGAN~\cite{choi2018stargan}, which contains two phases. In the forward phase, we first randomly generate the target-modality label $c'$, which is the one-hot encoding of the modalities, such as FLAIR, T1Gd and T2. We expand $c'$ to the size of the input image and perform depth-wise concatenation between expanded $c'$ and an arbitrary known modality image $x$. Given the input ($x$, $c'$), $G_{trans}$ learns to translate $x$ to target-modality image $x'$, $G_{trans}(x, c') \to x'$. Meanwhile, $G_{seg}$ takes $x$ as input and outputs the segmentation map $G_{seg}(x)$. In the backward phase, $G_{trans}$ takes fake image $x'$ and source-modality label $c$ as inputs and tries to recover the source-modality image $x$ by $G_{trans}(x', c)$. To preserve the tumor structure, an auxiliary shape-consistency loss~\cite{zhang2018translating} was added in the backward phase, which is the cross-entropy loss between $G_{seg}(x')$ and their manual annotations. 

In order to fool the discriminator $D$, $G_{trans}$ learns to modify image intensity relative to its neighbor tissues and keeps the brain structure unchanged. Utilizing these unchanged features of different modalities in $G_{trans}$ may have great potential to improve the segmentation performance. However, directly combining features from the translation task may not be an optimal method to get satisfactory segmentation results due to the features extracted by $G_{trans}$ which are not related to the segmentation task. 

To discard these features, we adopt attentional blocks at each upsampling step. The attentional blocks first generate attentional maps to highlight related and suppress unrelated features. Then the attentional blocks fuse the lower level feature maps from both streams with the attention maps. For more details, please refer to Section~\ref{attention section}. During inference, UAGAN conducts segmentation results given the testing images and their modality labels.
\subsection{Attentional Blocks} \label{attention section}
\begin{figure}
	\includegraphics[width=\textwidth]{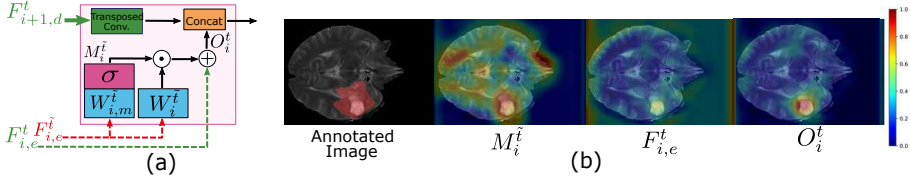}
	\caption{(a) The schematic illustration of attentional blocks (green for inputs from task $t$, red for another task $\tilde{t}$, best view in color). (b) Some examples of feature heatmaps\protect\footnotemark[1].} \label{atten}
\end{figure}

In the U-net architecture, the encoder captures multilevel features varying from low-level details to high-level semantic knowledge. The decoder combines the low-level and high-level features gradually to construct the final result. The features extracted by $G_{seg}$ are expected to be more related to the tumor. However, in the translation stream, $G_{trans}$ is trained to fool the discriminator $D$. Therefore, Unrelated information such as the contour and internal structures of the brain may be preserved in the translation encoder. To emphasize tumor-related features and suppress unrelated features , we propose attentional blocks at each upsampling step in the decoders. As presented in Fig.~\ref{atten}a, we denote the features extracted in encoder $e$ of task $t$ at $i$ level as $F^t_{i,e}$, where $i\in\{0,1,2,3\}$ corresponding to the levels of the feature maps, and $t\in\{trans,seg\}$ corresponding to the translation and segmentation tasks. Given the feature maps $F^{\tilde{t}}_{i,e}$ extracted from another task $\tilde{t}$, an attention map $M^{\tilde{t}}_i$ is first produced as follows:
\begin{equation}
M^{\tilde{t}}_i \gets \sigma(W^{\tilde{t}}_{i,m} \otimes F^{\tilde{t}}_{i,e})
\end{equation}
\noindent where $\otimes$ denotes the convolution operation, $\sigma$ is a sigmoid function and $W^{\tilde{t}}_{i,m}$ is the convolution mask.
\footnotetext[1]{Given a feature map with the size of $(c, h, w)$, we first apply summation at $c$ channel, and then normalize it for visualization.}
In order to reduce the information gap between $F^{\tilde{t}}_{i,e}$ and $F^t_{i,e}$, we apply a convolution to $F^{\tilde{t}}_{i,e}$ with parameters $W^{\tilde{t}}_i$ and then perform element-wise multiplication $\odot$ with the attention map $M^{\tilde{t}}_i$ to focus related information automatically. The fused outputs $O^t_i$ at level $i$ are produced as follows:
\begin{equation}
O^t_i \gets F^t_{i,e} + M^{\tilde{t}}_i \odot (W^{\tilde{t}}_i \otimes F^{\tilde{t}}_{i,e})
\end{equation}

The final outputs of the attentional block are the concatenation of $O^t_i$ and the upsampled $F^t_{i+1,d}$ extracted from the decoder $d$ of task $t$.
\section{Experimental Results}
\subsection{Materials}
Experiments were carried out on multimodal multisite MRI data released in the Brain Tumors Task of Medical Segmentation Decathlon\footnote[2]{\url{http://medicaldecathlon.com/}}. We selected 300 patients randomly and divided them into three disjoint partitions to form three independent datasets. Each dataset contained different modality images. Experiments were conducted on 2D slices of each dataset separately. In each dataset, 50 patients were used for training and 50 for testing. Each patient volume was consisted of 155 slices with the size of 240$\times$240 and a pixel size of 1$\times$1 mm in the direction of axis view. To form unpaired data in each dataset, we used only one of the modalities (T1Gd, FLAIR and T2) per patient randomly and applied z-score normalization to the volumes individually. To exclude the irrelevant regions, we neglected the slices of small brain tissues, cropped the center of slices of 200$\times$200, and then resized to 128$\times$128 due to the limitation of GPU memory. Finally, the average numbers of training and testing slices of each dataset are 5867 and 5861. To prevent overfitting, several data augmentation techniques (i.e., rotation, vertically flipped, horizontally flipped and scale) were applied on the fly during training. The whole framework was implemented in PyTorch, using a computer with the Intel i7 8700K CPU and an NVIDIA GTX 1080Ti GPU.
\subsection{Training Strategy}
\subsubsection{Loss Function.} To generate the indistinguishable target-modality images, the objective functions $\mathcal{L}_{D}$ to optimize the discriminator $D$ are the same as StarGAN~\cite{choi2018stargan}. However, the objective functions $\mathcal{L}_{G}$ to optimize the generator $G$ are significantly different because of uniting a new segmentation task:
\begin{equation}
\mathcal{L}_{G} = \mathcal{L}_{G_{trans}} + \mathcal{L}_{G_{seg}}
\end{equation}
\begin{equation}
\mathcal{L}_{G_{seg}} = \lambda_{seg} \mathcal{L}_{seg} + \lambda_{shape} \mathcal{L}_{shape}
\end{equation}
\noindent where the translation loss $\mathcal{L}_{G_{trans}}$ is defined in StarGAN~\cite{choi2018stargan} and the segmentation loss $\mathcal{L}_{G_{seg}}$ is defined as the summation of $\mathcal{L}_{seg}$ and $\mathcal{L}_{shape}$; $\mathcal{L}_{seg}$ and $\mathcal{L}_{shape}$ are the cross entropy losses in the forward and backward phase; $\lambda_{seg}$ and $\lambda_{shape}$ are the weights of the cross entropy losses. We set the weights ($\lambda_{seg}$, $\lambda_{shape}$) as 100 to emphasize the segmentation task. Other weights are set up the same as StarGAN~\cite{choi2018stargan}.
\subsubsection{Training Parameters.} In an end-to-end training manner, we updated the weights of all networks using the Adam optimizer with an initial learning rate of $1e^{-4}$, and the batch size of training was 8. All networks were trained up to 100 epochs, where the learning rate was fixed in the first 60 epochs and then linearly reduced to $1e^{-6}$. In the early phase, the synthetic images were blurry, so $\lambda_{shape}$ was set to 0 at the beginning and linearly increased to 100 at 60 epoch. In the end, we used the model trained at the 100th epoch to perform on testing data.
\subsection{Whole Tumor Segmentation Performance}
There are two types of multimodal segmentation methods, the unified and the non-unified model. In the unified model, all modality images are processed by a single stream, whereas in the non-unified model the processing is by the different streams. 
\begin{table}
	\centering
	\caption{Metric results of different models in multiple modalities (Best in bold).}\label{table1}
	\begin{tabular}{l|l|c|c|c|c|c}
		\toprule[2pt]
		Modality & Method & Dice(\%) & Precision(\%) & Sens(\%) & Spec(\%) & ASSD(mm) \\
		\toprule[2pt]
		\multirow{7}*{T1Gd} & Joint$\dagger$ & 53.77$\pm$5.42 & 56.19$\pm$1.59 & 58.38$\pm$8.29 & 99.13$\pm$0.06 &  8.81$\pm$1.10 \\
		& Multi V4~\cite{valindria2018multi} &57.41$\pm$0.69 & 54.18$\pm$1.07 & \bfseries{67.14$\pm$3.82} & 98.84$\pm$0.12 & 8.82$\pm$0.57 \\
		& Individual~\cite{ronneberger2015u} & \bfseries{57.79$\pm$0.28} & 56.82$\pm$3.56 & 65.01$\pm$4.18 & 98.99$\pm$0.18 & 8.14$\pm$0.44 \\
		\cline{2-7}
		& UAGAN-fuse$\dagger$ & 54.95$\pm$3.93 & 64.49$\pm$2.40 & 53.67$\pm$6.12 & 99.50$\pm$0.06 & \bfseries{7.81$\pm$1.41} \\
		& UAGAN-trans$\dagger$ & 53.53$\pm$4.34 & 65.40$\pm$1.19 & 50.21$\pm$5.94 & 99.58$\pm$0.03 & 8.11$\pm$0.42 \\
		& UAGAN-atten$\dagger$ &  53.87$\pm$5.14 & \bfseries{66.19$\pm$3.54} & 50.65$\pm$7.59 & \bfseries{99.59$\pm$0.10} & 8.41$\pm$1.52\\
		& UAGAN$\dagger$ & 53.61$\pm$5.50 & 66.11$\pm$3.02 & 51.14$\pm$7.66 & 99.57$\pm$0.08 & 8.46$\pm$1.09 \\
		\midrule[2pt]
		\multirow{7}*{FLAIR} & Joint$\dagger$ & 78.17$\pm$4.13 & 71.74$\pm$5.66 & 88.76$\pm$2.09 & 99.18$\pm$0.22 & 3.61$\pm$0.56 \\
		& Multi V4~\cite{valindria2018multi} & 79.56$\pm$1.97 & 73.78$\pm$3.62 & \bfseries{88.86$\pm$0.91} & 99.25$\pm$0.15 & 3.40$\pm$0.19 \\
		& Individual~\cite{ronneberger2015u} & 80.22$\pm$1.15 & 75.17$\pm$3.45 & 88.40$\pm$0.67 & 99.31$\pm$0.16 & 3.14$\pm$0.23 \\
		\cline{2-7}
		& UAGAN-fuse$\dagger$& 80.67$\pm$3.14 & 79.00$\pm$4.93 & 85.32$\pm$2.99 & 99.47$\pm$0.12 & 2.84$\pm$0.51 \\
		& UAGAN-trans$\dagger$& 80.67$\pm$3.73 & 80.55$\pm$4.57 & 84.11$\pm$3.16 & \bfseries{99.51$\pm$0.13} & 2.77$\pm$0.47 \\
		& UAGAN-atten$\dagger$& 81.38$\pm$3.14 & 79.93$\pm$4.93 & 86.08$\pm$2.94 & 99.47$\pm$0.13 & 2.64$\pm$0.40 \\
		& UAGAN$\dagger$ & \bfseries{81.55$\pm$2.96} & \bfseries{81.20$\pm$4.11} & 84.98$\pm$1.92 & \bfseries{99.51$\pm$0.13} & \bfseries{2.53$\pm$0.29} \\
		\midrule[2pt]
		\multirow{7}*{T2} & Joint$\dagger$ & 73.08$\pm$2.30 & 69.76$\pm$6.57 & \bfseries{81.55$\pm$1.60} & 99.28$\pm$0.14 & 4.43$\pm$0.78 \\
		& Multi V4~\cite{valindria2018multi} & 73.07$\pm$3.19 & 69.15$\pm$8.51 & 81.40$\pm$4.08 & 99.28$\pm$0.14 & 4.51$\pm$0.97 \\
		& Individual~\cite{ronneberger2015u} & 72.29$\pm$1.16 & 70.07$\pm$7.73 & 79.55$\pm$5.38 & 99.28$\pm$0.18 & 4.59$\pm$0.46 \\
		\cline{2-7}
		& UAGAN-fuse$\dagger$ & 75.32$\pm$2.12 & 78.94$\pm$4.77 & 77.03$\pm$0.73 & 99.59$\pm$0.10 & 3.65$\pm$0.39 \\
		& UAGAN-trans$\dagger$ & 75.98$\pm$1.61 & 80.13$\pm$5.98 & 76.51$\pm$2.15 & 99.60$\pm$0.10 & 3.59$\pm$0.44 \\
		& UAGAN-atten$\dagger$ & 75.05$\pm$3.74 & 79.24$\pm$4.65 & 76.08$\pm$2.33 & 99.58$\pm$0.06 & 3.65$\pm$0.56 \\
		& UAGAN$\dagger$ & \bfseries{76.54$\pm$2.91} & \bfseries{80.16$\pm$3.52} & 77.45$\pm$1.90 & \bfseries{99.60$\pm$0.08} & \bfseries{3.44$\pm$0.63} \\
		\bottomrule[2pt]
	\end{tabular}
\end{table}
As a baseline, we trained U-nets corresponding to each modality individually and tested them only on the corresponding modality (Individual). To compare with the unified model, we trained a U-net for all modalities (Joint). To compare with the same non-unified model, we trained a multistream model (Multi V4)~\cite{valindria2018multi} and changed the backbone from FCN to U-net for a fair comparison. Ablation studies were conducted by UAGAN, including only sharing the last blocks of encoders without fusing features from different streams (UAGAN-fuse) and simply adding the features of both streams without attention (UAGAN-atten). To verify the effect of the translation stream, we also replaced the translation task with the reconstruction task and kept the attention blocks unchanged (denoted as UAGAN-trans).

We employed five metrics to evaluate the performance of whole brain tumor segmentation, including Dice score (Dice), Precision, Sensitivity (Sens), Specificity (Spec) and Average Symmetric Surface Distance (ASSD). Because all the models were trained on 2D slices, we concatenated the slices of the same patient, and all the metrics were performed on 3D volumes.  A better model will have higher Dice, Sens, Spec, Precision and lower ASSD.
\begin{figure}
	\includegraphics[width=\textwidth]{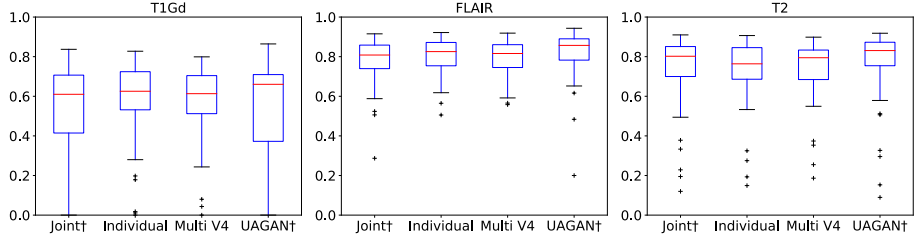}
	\caption{Box plot of different modalities tumor segmentation: T1Gd, FLAIR, and T2. Different models on x-axis and Dice scores on y-axis. } \label{box}
\end{figure}

\begin{figure}
	\includegraphics[width=\textwidth]{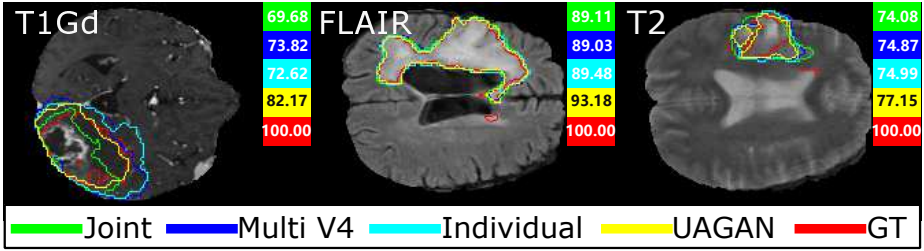}
	\caption{Visual comparison of the whole brain tumor segmentation results. Right side of the images: dice scores of the methods mentioned in Table \ref{table1}. Best view in color.} \label{vis}
\end{figure}

Experiments were conducted with three disjoint unpaired datasets, and the results are shown in Table~\ref{table1}. We compared the metrics in three modalities (T1Gd, FLAIR and T2). The symbol $\dagger$ denotes the unified model.
The boxplot in Fig. \ref{box} from all test cases shows the performance of different models in terms of Dice scores. 
Some visual segmentation results are shown in Fig. \ref{vis}.
Our methods outperformed the unified model and performed better than the non-unified models except some cases in T1Gd. The results revealed that similar modalities such as FLAIR and T2 could improve the performance of segmentation with each other.
\section{Conclusion}
In this work, we propose a novel two-stream unified attentional generative adversarial network (UAGAN) for multimodal unpaired medical image segmentation. Our framework is flexible, and can use on more than two modalities due to the unified structure and has fewer parameters than the same non-unified model. To capture the modality-invariant features beneficial to segmentation, we fuse the features from both segmentation and translation streams. Furthermore, feature recalibration is performed with attentional blocks to emphasize useful features. Experiments on brain tumor segmentation indicate that our framework achieved better performance in most cases. The more similar the modalities are, the more significant the effect will be, such as FLAIR and T2. Our proposed framework can alleviate the problem of paired multimodal medical image scarcity. In the future, the framework will be applied to other biomedical image segmentation tasks such as multimodal abdominal organ segmentation.

\subsubsection{Acknowledgments.}This work is supported by the National Natural Science Foundation of China (61402181, 61502174), the Natural Science Foundation of Guangdong Province (2015A030313215, 2017A030313358, 2017A030313355), the Science and Technology Planning Project of Guangdong Province (2016A040403046), the Guangzhou Science and Technology Planning Project (201704030051).
%
%
%
\bibliographystyle{splncs04}
\bibliography{ref643}

\end{document}